# Exploring Distributed Control with the NK Model

Larry Bull

Computer Science Research Centre

University of the West of England, Bristol BS16 1QY UK

Larry.Bull@uwe.ac.uk

**Abstract**

The NK model has been used widely to explore aspects of natural evolution and complex systems. This paper introduces a modified form of the NK model for exploring distributed control in complex systems such as organisations, social networks, collective robotics, etc. Initial results show how varying the size and underlying functional structure of a given system affects the performance of different distributed control structures and decision making, including within dynamically formed structures and those with differing numbers of control nodes.

**Introduction**

Kauffman and Levin [7] introduced the NK model to allow the systematic study of various aspects of organisms evolving on fitness landscapes, and it has since been applied to a number of biological phenomena (eg, see [2]). Given its abstract nature, the model has also been used widely within complexity science, particularly around organization and management studies (eg, see [8]). A distributed version of the NK model, termed the NKCS model [6] was later introduced to explore coevolution, that is, the evolutionary dynamics of ecosystems containing multiple species. Again, versions of the model have been applied to non-biological systems, particularly spatially extended versions, for example in "patches" to receiver-based communication optimisation (see [5]).

Distributed control is applicable across the spectrum of complex systems, from spatially extended plant control (eg, see [9]), to collective robotics (eg, see [3]), to environmental science (eg, see [1]). Typically, the underlying functional structure which exists between the constituent parts of the system is only partially known or understood and hence the utility of different distributed control structures is unclear. That is, the application of any given distributed control structure may result in sub-optimal performance. This paper introduces a new version of the NK model through which to systematically explore the general properties of such systems – the NKD model. Initial results presented here show the relationship between the underlying functional structure and the optimal distributed control structure, that equivalent performance to global control exists within a significant proportion of the attribute space of the model, and that temporally dynamic control structures can prove beneficial.

**The NK Model**

In the standard NK model, the features of the genome/system are specified by two parameters: $N$, the length of the genome; and $K$, the number of genes that has an effect on the fitness contribution of each (binary) gene. Thus increasing $K$ with respect to $N$ increases the epistatic linkage, increasing the ruggedness of the fitness/problem landscape. The increase in epistasis increases the number of optima, increases the steepness of their sides, and decreases their correlation. The model assumes all intragenome interactions are so complex that it is only appropriate to assign random values to their effects on fitness. Therefore for each of the possible $K$ interactions a table of $2^{(K+1)}$ fitnesses is created for each gene with all entries in the range 0.0 to 1.0, such that there is one fitness for each combination of traits (Figure 1). The fitness contribution of each gene is found from its table. These fitnesses are then summed and normalized by $N$ to give the selective fitness of the total genome.

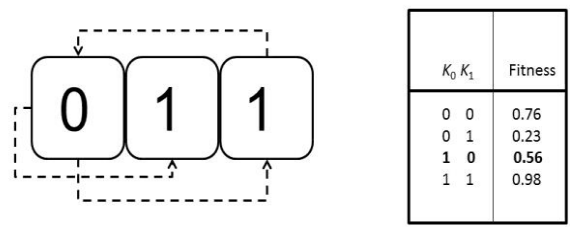

Figure 1. An example NK model ($N$=3, $K$=1) showing how the fitness contribution of each gene depends on $K$ random genes (left). Therefore there are $2^{(K+1)}$ possible allele combinations per gene, each of which is assigned a random fitness. Each gene of the genome has such a table created for it (right, centre gene shown). Total fitness is the normalized sum of these values.

Kauffman [4] used a mutation-based hill-climbing algorithm, where the single point in the fitness space is said to represent a converged species, to examine the properties and evolutionary dynamics of the NK model. That is, the population is of size one and a species evolves by making a random change to one randomly chosen gene per generation. The "population" is said to move to the genetic configuration of the mutated individual if its fitness is greater than the fitness of the current individual; the rate of supply of mutants is seen as slow compared to the actions of selection. Ties are broken at random. Figure 2 shows example results. All results reported in this paper are the average of 10 runs (random start points) on each of 10 NK functions, ie, 100 runs, for 5000 generations. Here $0 \leq K \leq 15$, for $N=20$ and $N=100$.

Figure 2 shows examples of the general properties of adaptation on such rugged fitness landscapes identified by Kauffman (eg, [4]), including a "complexity catastrophe" as $K \rightarrow N$. When $K=0$ all genes make an independent contribution to the overall fitness and, since fitness values are drawn at random between 0.0 and 1.0, order statistics show the average value of the fit allele should be 0.66. Hence a single, global optimum exists in the landscape of fitness 0.66, regardless of the value of $N$. At low levels of $K$ ($0<K<8$), the landscape buckles up and becomes more rugged, with an increasing number of peaks at higher fitness levels, regardless of $N$. Thereafter the increasing complexity of constraints between genes means the height of peaks typically found begin to fall as $K$ increases relative to $N$: for large $N$, the central limit theorem suggests reachable optima will have a mean fitness of 0.5 as $K \rightarrow N$. Figure 2 shows how the optima found when $K>6$ are significantly lower for $N=20$ compared to those for $N=100$ (T-test, $p<0.05$).

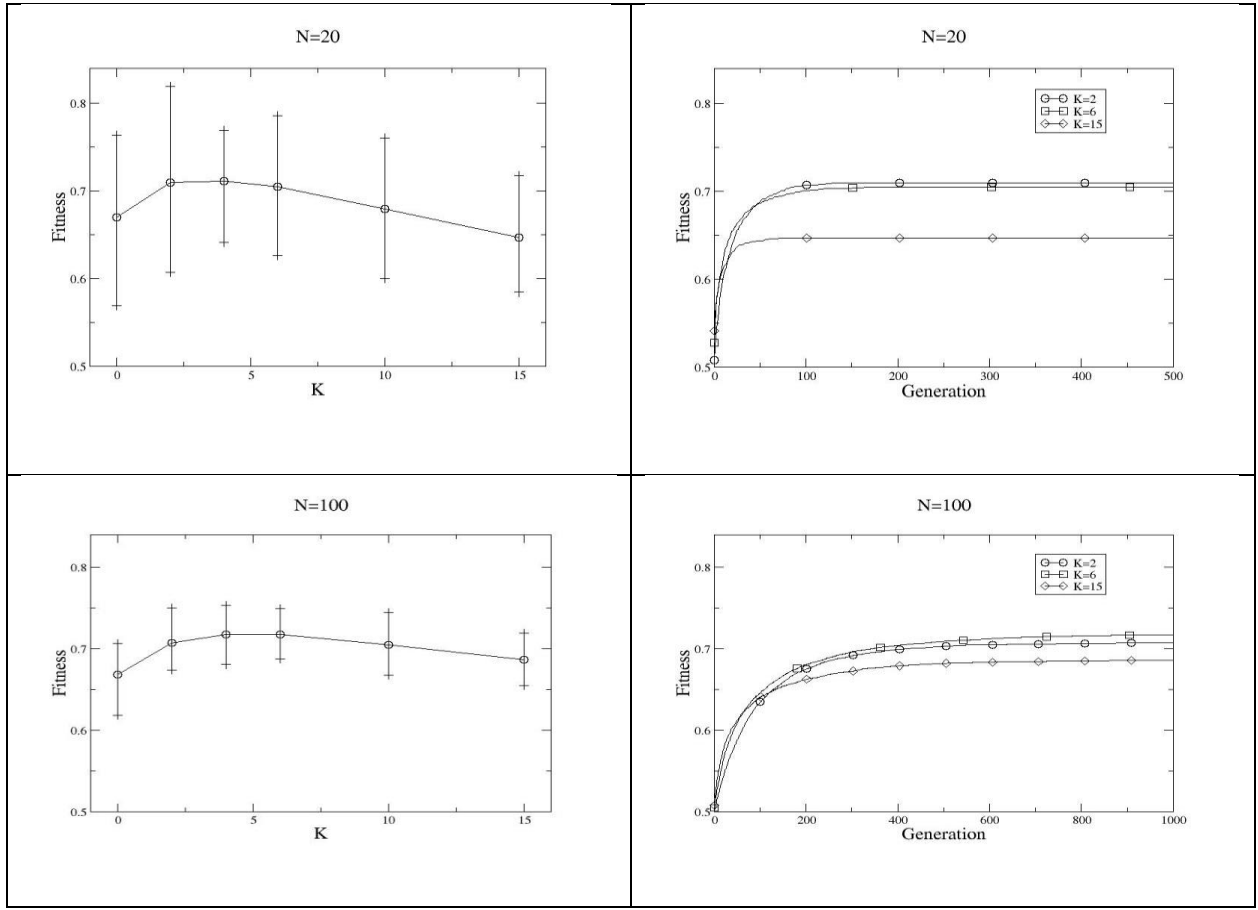

Figure 2. Showing typical behaviour and the fitness reached after 5000 generations on landscapes of varying ruggedness (*K*) and length (*N*). Error bars show min and max values.

**The NKCS Model**

Kauffman and Johnsen [6] extended the NK model and introduced the NKCS model to enable the study of various aspects of *co*evolution. At an abstract level coevolution can be considered as the coupling together of the fitness landscapes of the interacting species. Hence the adaptive moves made by one species in its fitness landscape causes deformations in the fitness landscapes of its coupled partners. In their model, each gene is also said to depend upon *C* randomly chosen traits in each of the other *S* species with which it interacts. The adaptive moves by one species may deform the fitness landscape(s) of its partner(s). Altering $C$, with respect to $N$, changes how dramatically adaptive moves by each species deform the landscape(s) of its partner(s). Again, for each of the possible $K+(SxC)$ interactions, a table of $2^{(K+(SxC)+1)}$ fitnesses is created for each gene, with all entries in the range 0.0 to 1.0, such that there is one fitness for each combination of traits. Such tables are created for each species (Figure 3, the reader is referred to [4] for full details).

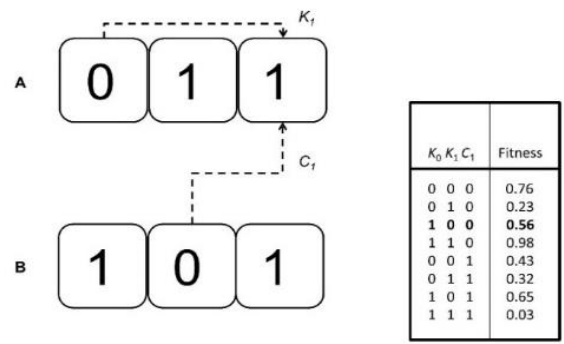

Figure 3. The NKCS model: Each gene is connected to $K$ randomly chosen local genes and to $C$ randomly chosen genes in each of the $S$ other species. A random fitness is assigned to each possible set of combinations of genes. These are normalised by $N$ to give the fitness of the genome. Connections and table shown for one gene in one species for clarity. $N=3$, $K=1$, $C=1$, $S=1$ here.

Figure 4 shows example results for one of two coevolving species ($S=1$) where the parameters of each are the same and hence behaviour is symmetrical. Again, as with the NK model, a species is said to be converged, $0 \leq K \leq 10$, $1 \leq C \leq 5$, for $N=20$ and $N=100$.

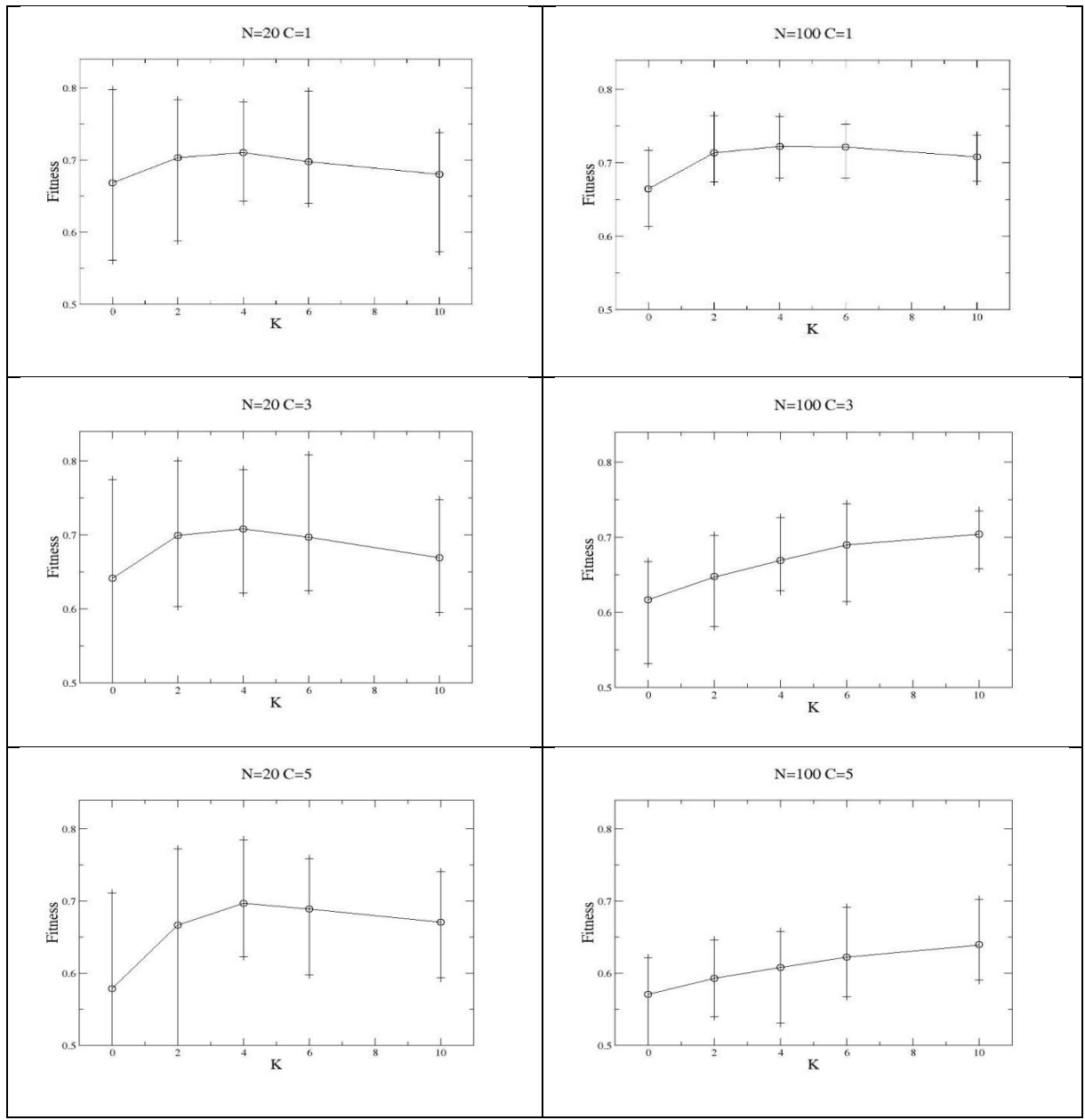

Figure 4. Showing the fitness reached after 5000 generations on landscapes of varying ruggedness (*K*), coupling (*C*), and length (*N*).

When *C*=1, Figure 4 shows examples of the general properties of adaptation on such fitness landscapes identified in the NK model, ie, when *C*=0, still hold, including the height of peaks typically found begins to fall as *K* increases relative to *N*. Figure 4 further shows how increasing the degree of connectedness (*C*) between the two landscapes causes fitness levels to fall significantly (T-test, $p<0.05$) when *C*≥*K* for *N*=20. That is, as *K*→*N* a high number of peaks of similar height typically exist in each of the fitness landscapes and so the effects of switching between them under the influence of $C$ is reduced since each landscape is very similar. Note this change in behaviour around $C=K$ was suggested as significant in [4] where $N=24$ only was used throughout. However, Figure 4 also shows how with $N=100$ fitness *always* falls significantly with increasing $C$ (T-test, $p<0.05$), regardless of $K$. That is, it might be concluded that more complex organisms ($>N$) appear more sensitive to landscape coupling ($>C$).

**The NKD Model**

The traditional NK model as described above can be seen to exercise global control in that any effect upon fitness caused by a mutation/change considers all $N$ genes/elements in the decision as to whether to accept that mutation/change. The standard NKCS model, as described above, can be seen to include a particular form of decentralised control in the decision process since each of the individual $S+1$ NK models accepts a mutation/change based upon the effect within its own model. That is, an NKCS model can be viewed as dividing a traditional NK model into $S+1$ non-overlapping components, each making local decisions (Figure 5).

The NKD model is introduced as a generalisation of the NKCS model to enable the exploration of distributed control structures within the space of NK models. Here each gene/element in the traditional NK model is extended to include $D$ connections to other genes/elements (Figure 5). In the simple case these are assigned at random. When a mutation/change is made to a given gene/element, the decision as whether to accept that change is based upon the effect on fitness of the set of $D+1$ genes/elements. Hence the traditional case exists when $D=(N-1)$ and the NKCS model exists if $D$ is assigned within blocks of mutually connected genes/elements. The networks of decision making will typical overlap in the NKD model (eg, see [10] for early discussions).

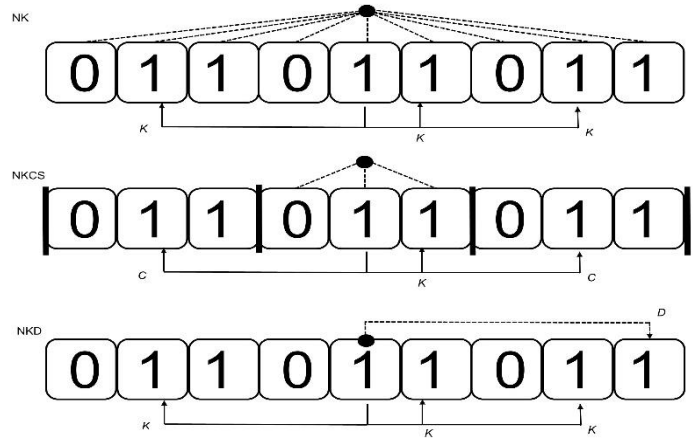

Figure 5. Showing the centralised, global control of the traditional NK model ($N=9$, $K=3$), the distributed non-overlapping control structure of the NKCS model ($N=3$, $K=1$, $C=1$, $S=2$), and the fully distributed control structure of the NKD model ($N=9$, $K=3$, $D=1$). Connections shown only for one gene/element for clarity.

As Figure 6 shows, with $N=20$, the control structure has no effect when $K=0$ since each gene/component makes an independent contribution to the overall fitness, whereas fitness is the same as $D=(N-1)$ when $D \geq 12$ for $0<K<10$ (T-test, $p<0.05$). For $K>6$ this is true for $D \geq 8$ and when $K=15$, ie, as $K \rightarrow N$, the highest fitness seen is at $D=8$ (T-test, $p<0.05$). Figure 7 shows similar results for $N=100$, although relative fitness is much worse for $D<60$ when $2<K<10$. For higher $K$, fitness remains lower than at $D=N-1$ when $D<80$ (T-test, $p<0.05$).

Hence whilst the best control structure varies with both $N$ and $K$, it is never only at $D=(N-1)$, ie, many distributed structures exist which give equivalently good performance to the traditional centralised scheme. It can be noted that results remain unchanged if the first $K$ connections of the $D$ connections (or up to $K$ when $D<K$) are the same (not shown). That is, this remains true regardless of any significant degree of correlation between the functional and control structures of the system.

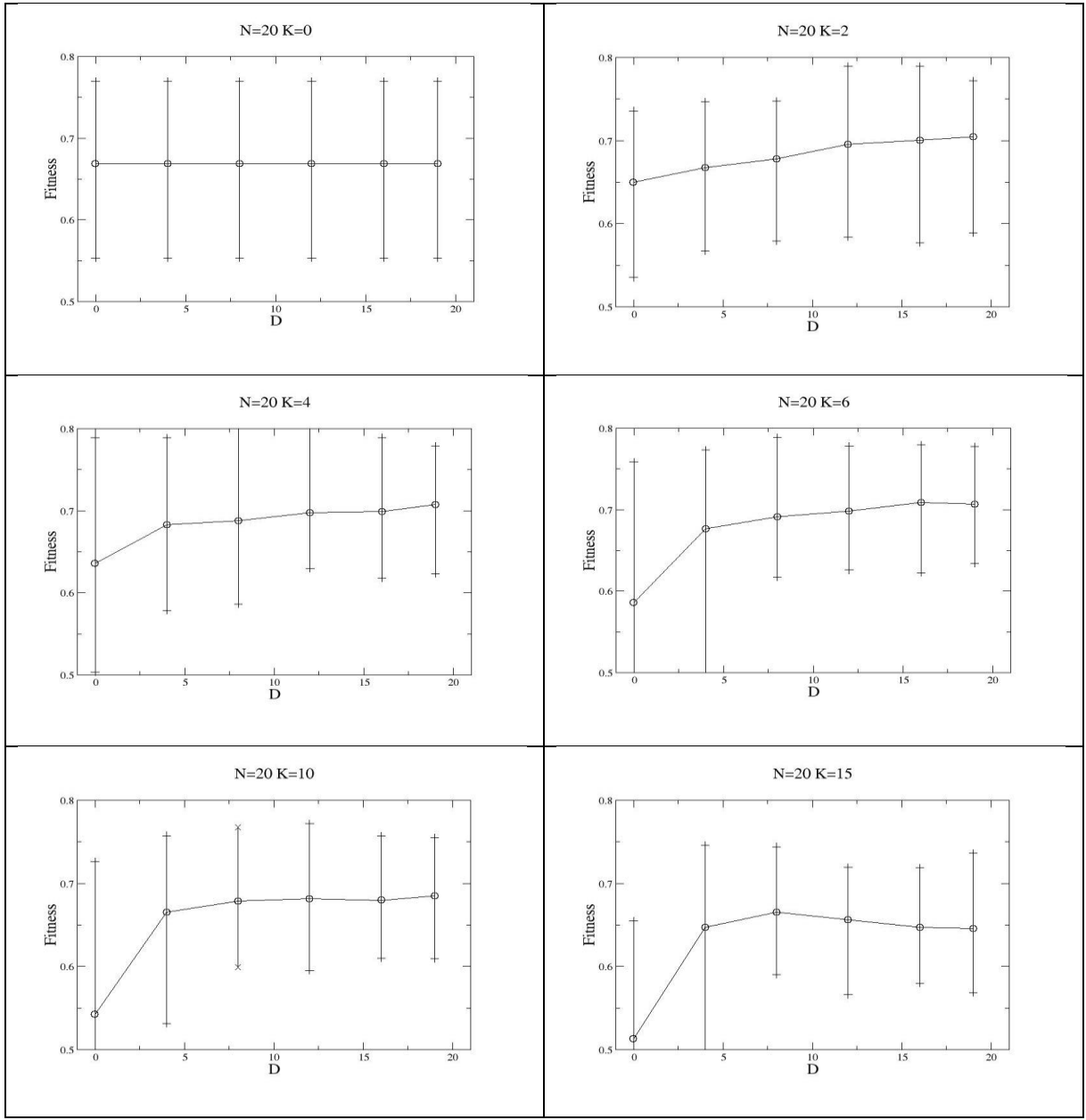

Figure 6. Showing the fitness reached after 5000 generations on landscapes of varying ruggedness ($K$) and degrees of control ($D$) with $N$=20.

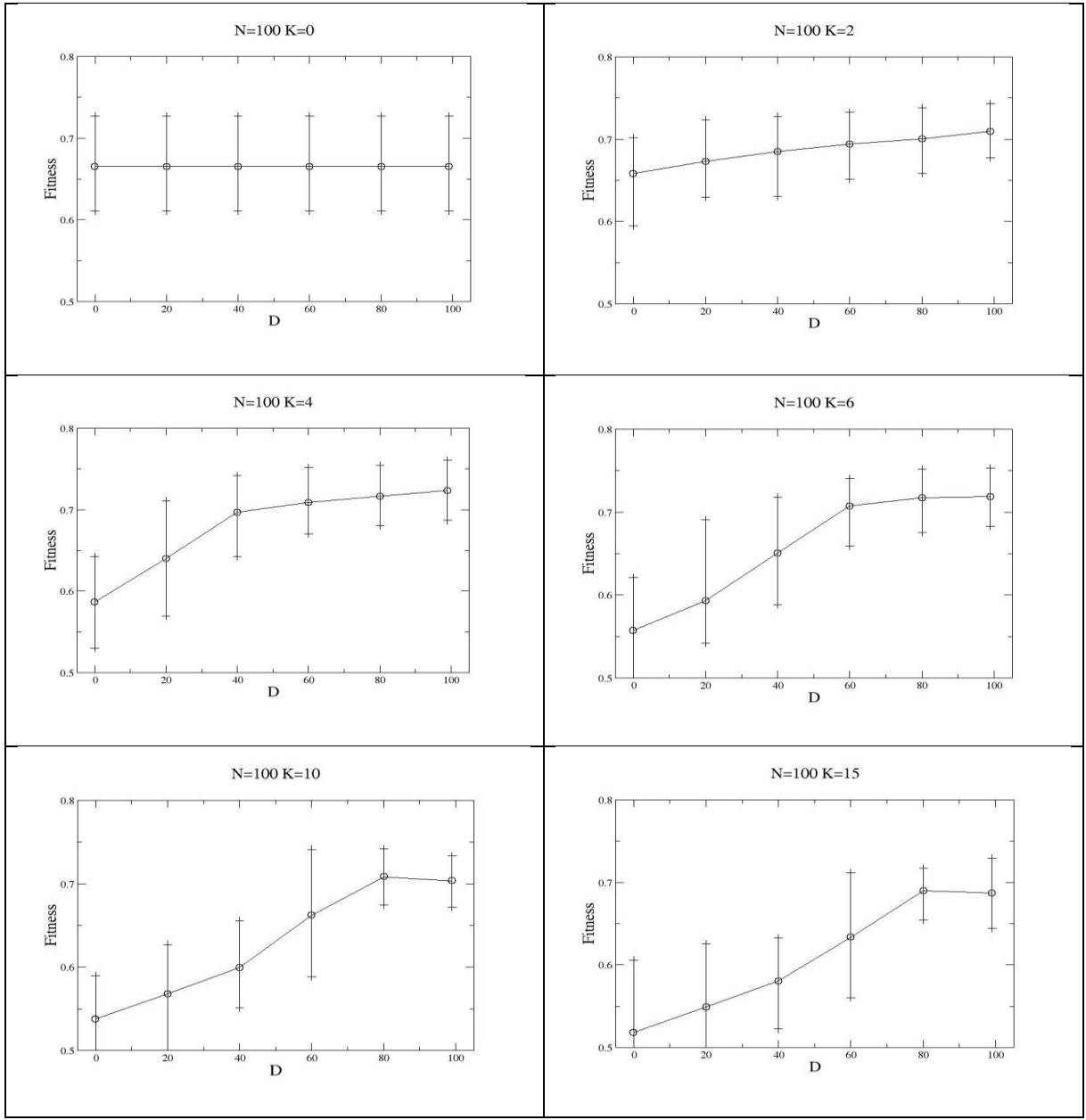

Figure 7. Showing the fitness reached after 5000 generations on landscapes of varying ruggedness ($K$) and degrees of control ($D$) with $N$=100.

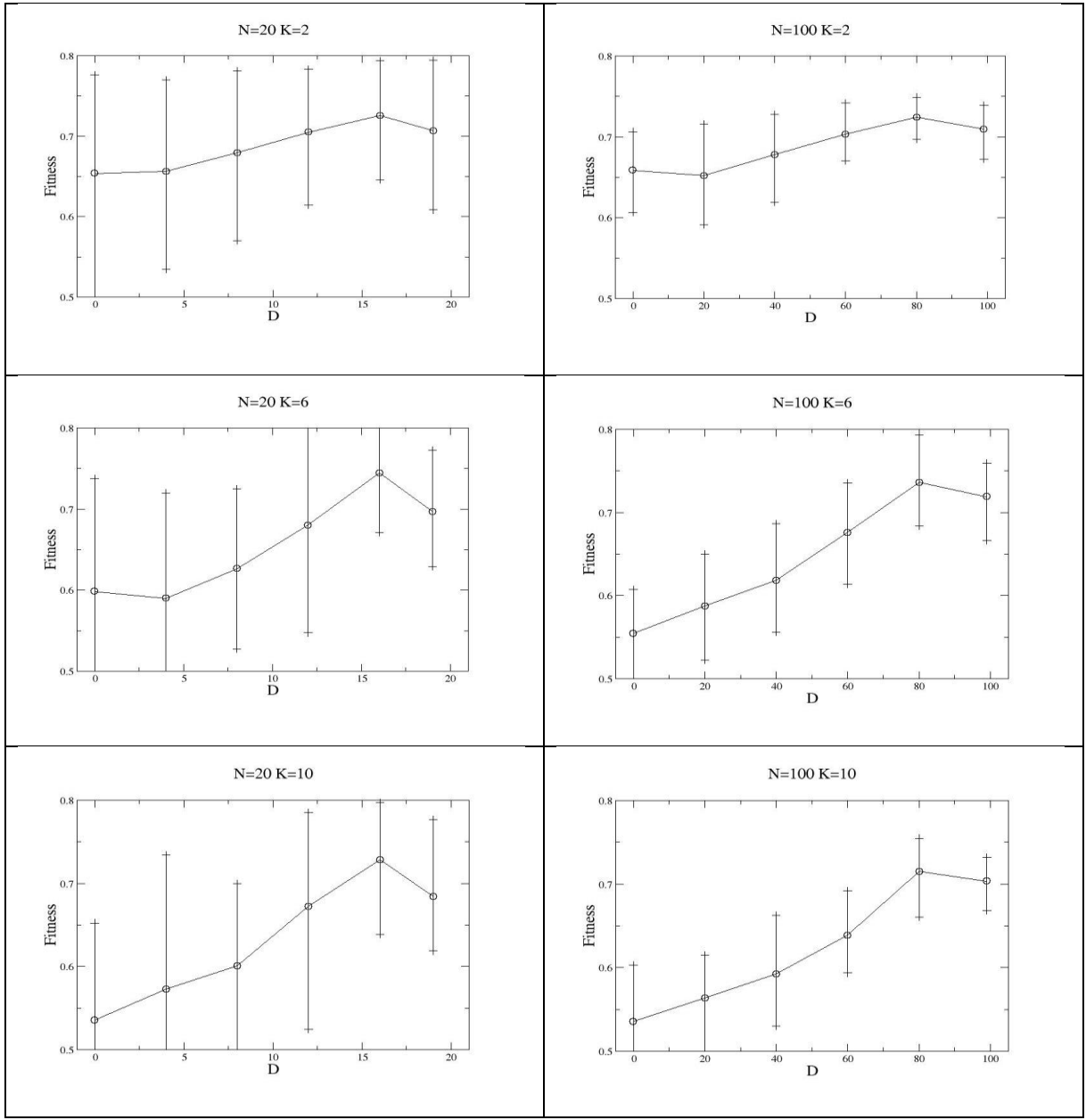

Figure 8. Showing the fitness reached after 5000 generations on landscapes of varying ruggedness (*K*), degree of control (*D*), and length (*N*) when the control structure is assigned at random per generation.

It can be noted that with larger *N*, fitness is significantly reduced for all *K*>0 and *D*>0 when more than one gene/element is altered per generation, even with only two changes (T-test, *p*<0.05) (not shown). Whereas smaller *N* systems remain insignificantly altered, even for four changes (T-test, *p*≥0.05) (not shown).

In some distributed system scenarios the topology of the control structure can vary temporally, eg, due to changes in geographic location in systems containing mobile elements. Figure 8 shows example results when the topology of the $D$ connections of each gene/element is randomly (re)created on each generation. There is a significant drop in fitness for low $D$ compared to $D=(N-1)$ in all cases. Perhaps most surprisingly, optimal performance is typically seen when $D \approx 0.8N$, regardless of $N$ (T-test, $p<0.05$). That is, distributed control can be beneficial when the structure is varied temporally in comparison to the traditional global control scheme. A similar idea has recently been explored within coevolutionary optimization [12].

Finally, the number of genes/elements which use a control configuration, ie, $D>0$, can be specified. That is, $d$ randomly chosen genes/elements are assigned $D$ connections with all others using $D=0$. This is loosely related to the idea of structural control and the identification of a subset of nodes to receive a given stimulus to shape global system behaviour (eg, see [11]). Figure 9 shows how, regardless of $K$ and $D$, $d \geq 12$ typically gives the same performance as $d=N$ with $N=20$ (T-test, $p<0.05$). However, Figure 10 shows how with $N=100$ performance is typically worse than $d=N$ for $K>2$, regardless of $D$ (T-test, $p<0.05$).

**Conclusion**

Distributed control is ubiquitous in natural and social systems. This paper has introduced an extension to a well-known abstract model used to study complex systems to explore distributed control. The results indicate that system size ($N$), functional connectivity ($K$), control connectivity ($D$) and degree ($d$) can all affect overall system performance. However, equivalent or better performance in comparison to a single, global point of control is consistently found across the parameter space of the model. Future work will consider the use of hierarchical control.

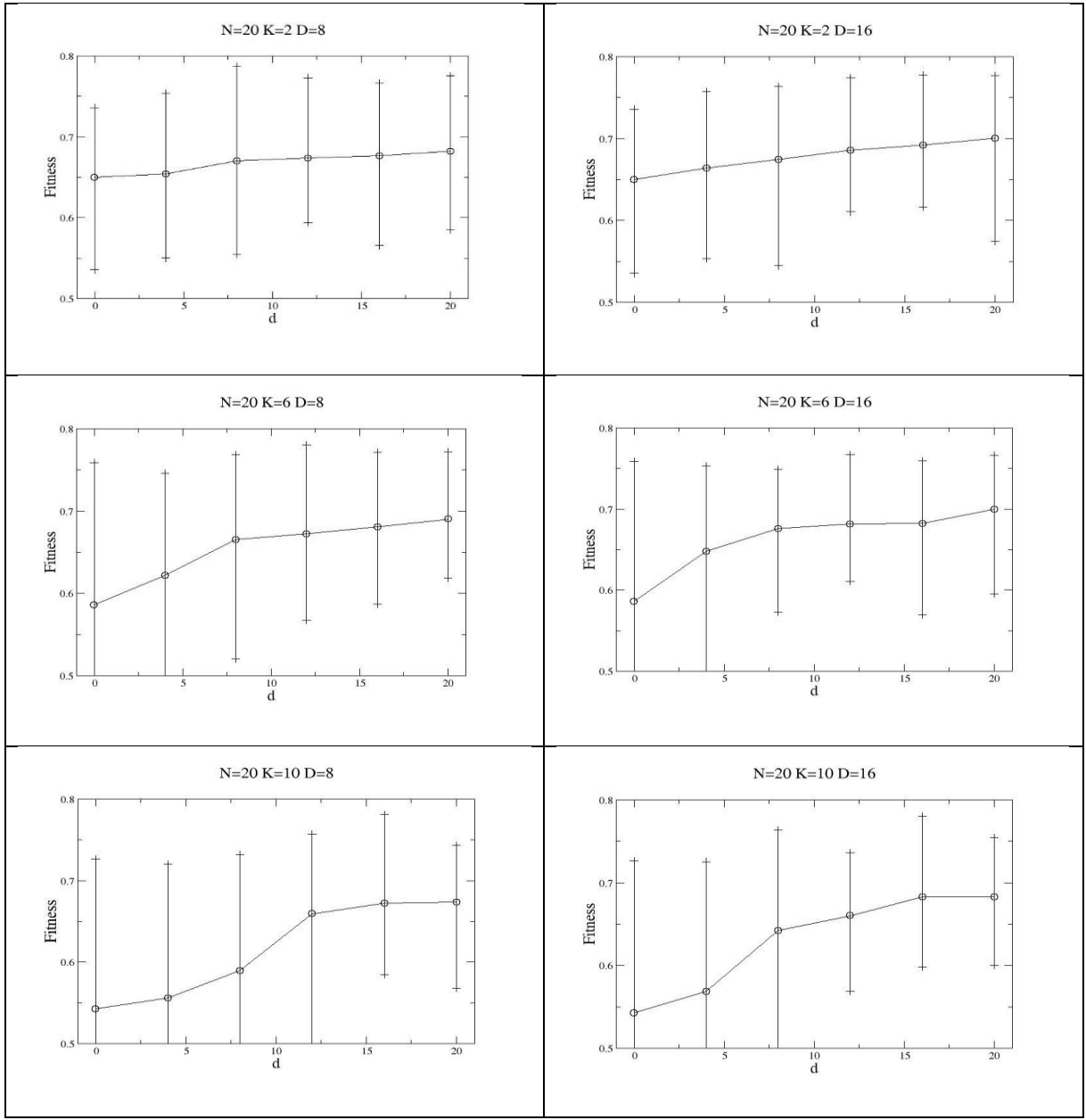

Figure 9. Showing the fitness reached after 5000 generations on landscapes of varying ruggedness (*K*), degrees of control (*D*), and control nodes (*d*) with *N*=20.

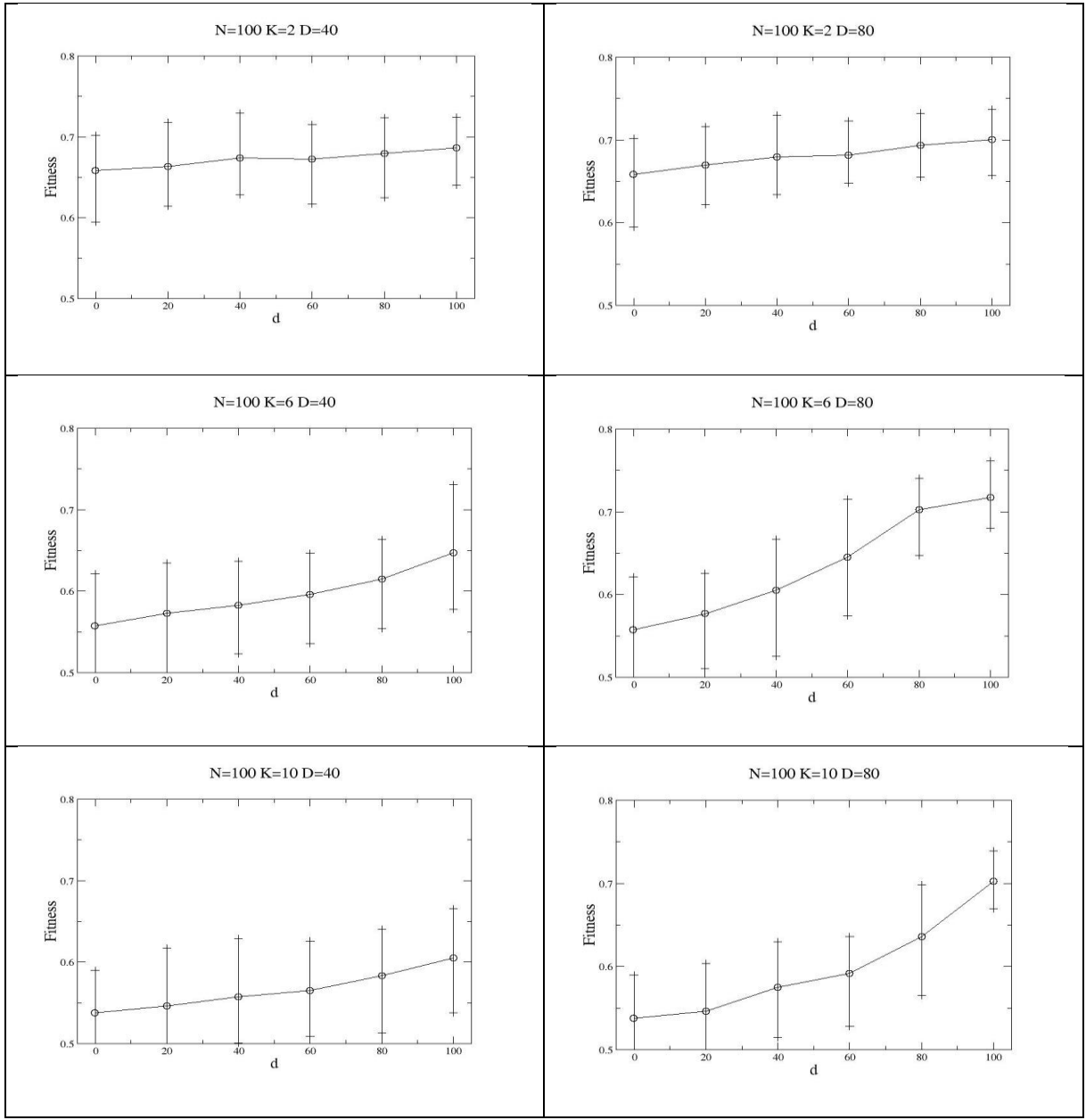

Figure 10. Showing the fitness reached after 5000 generations on landscapes of varying ruggedness (*K*), degrees of control (*D*), and control nodes (*d*) with *N*=100.